\documentclass[10pt,twocolumn,letterpaper]{article}

\usepackage{iccv}
\usepackage{times}
\usepackage{epsfig}
\usepackage{graphicx}
\usepackage{amsmath}
\usepackage{amssymb}

\usepackage{array}
\usepackage{multirow}
\usepackage{color}
\usepackage{xcolor}
\usepackage{verbatim}
\usepackage{booktabs}
\usepackage{bigstrut}
\usepackage{amsthm}

\def \R {\mathbb{R}}
\def \x {\mathbf{x}}
\def \w {\mathbf{w}}
\def \z {\mathbf{z}}
\def \LL {\mathcal{L}}

\newtheorem{thm}{Theorem}
\newtheorem{prop}{Proposition}
\newtheorem{cor}{Corollary}

\usepackage[accsupp]{axessibility}  

\usepackage[breaklinks=true,bookmarks=false]{hyperref}

\iccvfinalcopy 

\def\iccvPaperID{10523} 

\ificcvfinal\fi

\begin{document}

\title{Improved Visual Fine-tuning with Natural Language Supervision}

\author{Junyang Wang$^1$\thanks{Work done during internship at DAMO Academy, Alibaba Group.}\quad Yuanhong Xu$^2$\quad Juhua Hu$^3$\quad Ming Yan$^2$\quad Jitao Sang$^{1,4}$\quad Qi Qian$^5$\thanks{Corresponding author}\\
$^1$ School of Computer and Information Technology \& Beijing Key Lab of Traffic Data Analysis and Mining,\\Beijing Jiaotong University, Beijing, China\\
$^2$ DAMO Academy, Alibaba Group, Hangzhou, China\\
$^3$ School of Engineering and Technology, University of Washington, Tacoma, WA 98402, USA\\
$^4$ Peng Cheng Lab, Shenzhen, China\\
$^5$ DAMO Academy, Alibaba Group, Bellevue, WA 98004, USA\\
{\tt\small \{junyangwang, jtsang\}@bjtu.edu.cn, \{yuanhong.xuyh, ym119608, qi.qian\}@alibaba-inc.com, juhuah@uw.edu}
}

\maketitle
\ificcvfinal\thispagestyle{empty}\fi

\begin{abstract}
   Fine-tuning a visual pre-trained model can leverage the semantic information from large-scale pre-training data and mitigate the over-fitting problem on downstream vision tasks with limited training examples. While the problem of catastrophic forgetting in pre-trained backbone has been extensively studied for fine-tuning, its potential bias from the corresponding pre-training task and data, attracts less attention. In this work, we investigate this problem by demonstrating that the obtained classifier after fine-tuning will be close to that induced by the pre-trained model. To reduce the bias in the classifier effectively, we introduce a reference distribution obtained from a fixed text classifier, which can help regularize the learned vision classifier. The proposed method, Text Supervised fine-tuning (TeS), is evaluated with diverse pre-trained vision models including ResNet and ViT, and text encoders including BERT and CLIP, on 11 downstream tasks. The consistent improvement with a clear margin over distinct scenarios confirms the effectiveness of our proposal. Code is available at \url{https://github.com/idstcv/TeS}.
\end{abstract}

\section{Introduction}

Fine-tuning a pre-trained visual deep model has become a prevalent paradigm for vision categorization~\cite{kornblith2019better}. By initializing the model with parameters pre-trained on a large-scale data set, fine-tuning can effectively transfer the semantic information from the pre-training data to diverse downstream tasks, which is essential to mitigate the over-fitting problem on data sets with limited training examples~\cite{HeGD19}.

\begin{figure}[t]
    \centering
    \includegraphics[width=0.45 \textwidth]{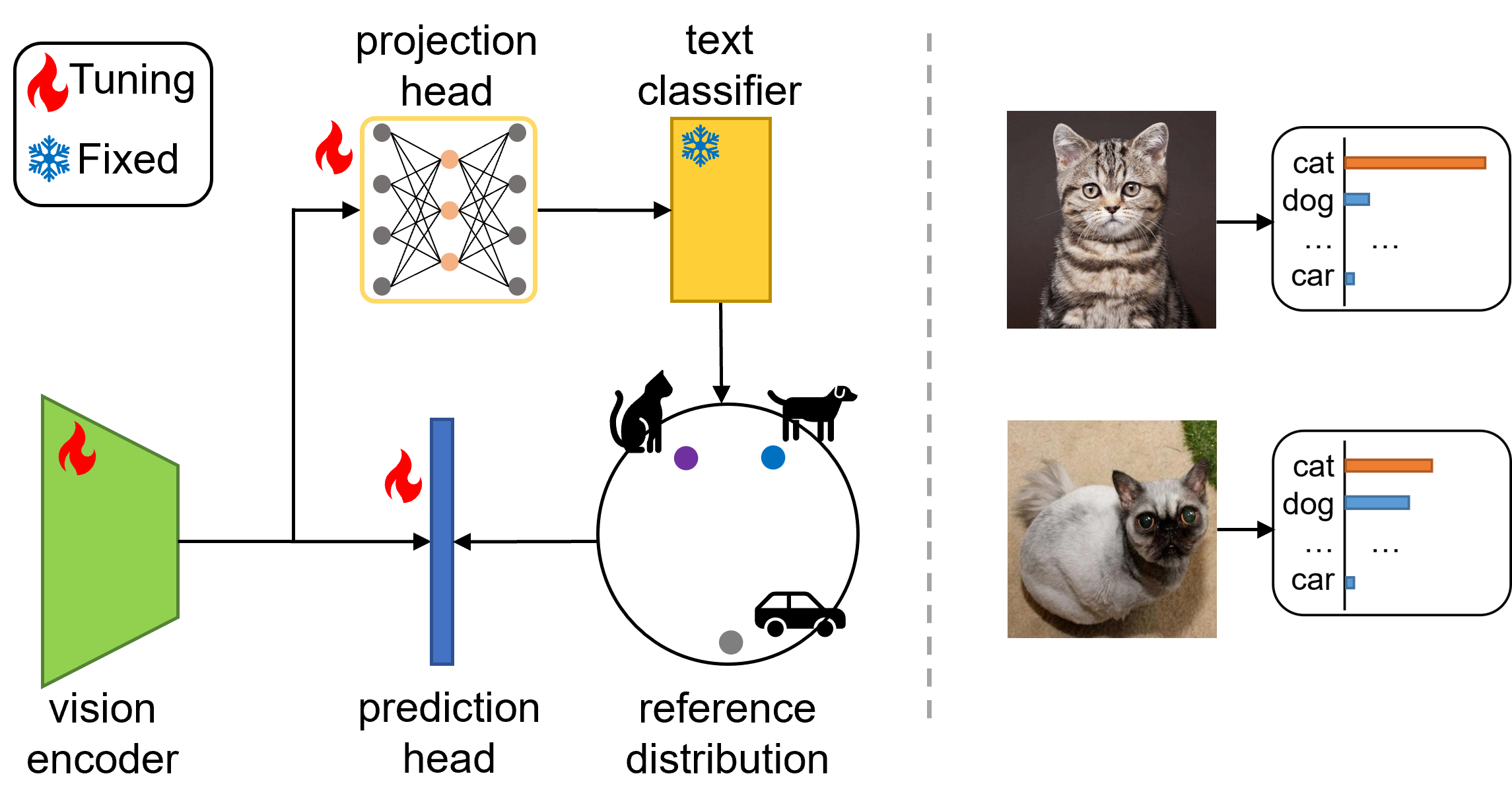}
    \caption{Illustration of the proposed method, TeS. During fine-tuning, the vision classifier is regularized by reference distributions defined with a fixed text classifier, which is obtained from class names. In TeS, diverse reference distributions can be obtained for different examples even from the same class (e.g., cats on the right side).}
    \label{fig:illu}
\end{figure}

While supervised pre-training methods~\cite{kornblith2019better} require full label information for the whole data set, the recent progress on self-supervised learning shows that an applicable pre-trained model can be obtained from unlabeled data~\cite{He0WXG20}. Consequently, a lot of research efforts have been devoted to exploring unsupervised pre-training, which can eliminate the labeling cost for a vast amount of examples. Many effective self-supervised methods have been developed, e.g., instance discrimination~\cite{ChenK0H20,He0WXG20}, cluster discrimination~\cite{CaronMMGBJ20,0001XHL022}, and masked image modeling~\cite{HeCXLDG22}. Compared to the supervised counterparts, fine-tuning pre-trained models from these unsupervised methods can achieve comparable or even better performance on downstream tasks.

With the success of pre-training, fine-tuning has thus attracted much attention to leverage the pre-trained model appropriately for downstream tasks. One main challenge for fine-tuning is the catastrophic forgetting in pre-trained backbone~\cite{ChenWFLW19}, that is, after sufficient learning with a large learning rate, the fine-tuned model becomes far away from the pre-trained model, incurring the over-fitting problem. The challenge has been extensively studied and can be tackled by simply setting a small learning rate for backbone or constraining the distance from the fine-tuned model to the pre-trained model explicitly~\cite{LiGD18,LiXWRLH19}.

The other challenge for fine-tuning is the potential bias existing in a pre-trained model. It should be noted that a pre-trained model is optimized by a specific pretext task using only the pre-training data. Therefore, if the data distribution of a target task is different from that of the pre-training data, the pre-trained model is biased to the pre-training data distribution~\cite{QianHL20}. This bias problem becomes more challenging in the scenario together with catastrophic forgetting, where the bias cannot be eliminated sufficiently with a constraint making the fine-tuned model and pre-trained model close. A recent work~\cite{liu2021improved} proposes to add a subset of pre-training data that is related to the target task for fine-tuning. The optimization with partial pre-training data can help preserve effective information in the pre-trained model to avoid over-fitting while reducing the bias for the target task. However, it requires the access of the pre-training data, which is infeasible in many real applications.

Recently, it has been found that side information from other related modalities can help visual pre-training. For example, CLIP~\cite{radford2021learning} pre-trains dual encoders by optimizing image-text pairs that align images with their corresponding text descriptions. The obtained model shows a strong zero-shot transfer performance, which can classify images using proxies consisting of text representations extracted from class names. The observation implies that text information is capable of guiding visual representation learning. However, the superior zero-shot performance highly depends on the paired vision and text encoders in pre-training, and thus is not applicable for an arbitrary vision encoder.

Inspired by vision-language pre-training~\cite{radford2021learning}, we introduce natural language supervision to fine-tuning, so as to mitigate the contradiction problem between the bias in pre-trained vision models and catastrophic forgetting in fine-tuning them. First, we show that without any side information, the classifier induced by the pre-trained vision model will be largely preserved after fine-tuning on the target data, which demonstrates the potential bias exiting in the conventional fine-tuning pipeline. To reduce the bias in the learned classifier, we propose to include text supervision as the reference information. Concretely, with a fixed pre-trained text encoder, a text classifier for the target task can be obtained by extracting proxies with class names. Given the text classifier, both class-level and instance-level reference distributions can be obtained for the target vision task. Then, the pre-trained vision models can be fine-tuned with the appropriate reference distributions for the vision classifier. Since the reference regularization is independent from the backbone, our method can reduce the bias without catastrophic forgetting. 

The proposed \textbf{Te}xt \textbf{S}upervised fine-tuning (TeS) method is illustrated in Fig.~\ref{fig:illu}. It should be noted that reference distributions are extracted from a fixed text encoder, and thus the overhead for the text supervision is negligible. The main contributions of this work can be summarized as follows.
\begin{itemize}
\item This work proposes to leverage text supervision from a fixed text encoder for fine-tuning an arbitrary pre-trained vision model. The text encoder can be pre-trained on a large corpus with rich context information, making it a good complement to vision tasks. 
\item With the classifier consisting of proxies from text encoder, we investigate the class-level and instance-level distributions for regularizing the fine-tuning of the vision classifier. By minimizing the derived cross entropy loss, the vision model can exploit the side information from the text supervision to reduce the bias from the pre-trained models.
\item Experiments on extensive downstream tasks demonstrate the effectiveness of text supervision for visual fine-tuning. In addition, the CLIP text encoder outperforms the BERT encoder~\cite{DevlinCLT19} and it implies that the text encoder pre-trained with vision-language pretext tasks is more appropriate for supervising visual tasks. 
\end{itemize}

\section{Related Work}

In this section, we briefly review visual pre-training and visual fine-tuning.

\subsection{Visual Pre-training}

A generic visual pre-training paradigm includes both supervised and self-supervised approaches. Supervised pre-training requires a large number of labeled data and can learn rich semantic information for fine-tuning~\cite{QianHL20,sutskever2013importance,zhu2023debiased}. 

To eliminate the cost of labeling, self-supervised learning is developed to obtain pre-trained models from unlabeled data. Many pretext tasks were proposed for effective learning, e.g., instance discrimination that considers each instance as an individual class and optimizes random augmentations from the same instance~\cite{chen2020simple,He0WXG20}, cluster discrimination that explores the relationship between different instances~\cite{CaronMMGBJ20,0001XHL022} and masked image modeling that leverages information within each image~\cite{he2022masked}. Moreover, \cite{zoph2020rethinking} demonstrates that self-supervised pre-training improves supervised pre-training with strong data augmentations. 

\subsection{Visual Fine-tuning}
After pre-training, fine-tuning aims to facilitate the downstream tasks by transferring semantic information in the pre-trained models~\cite{kornblith2019better}. One main challenge for fine-tuning is from catastrophic forgetting that the fine-tuned model over-fits target data and becomes far away from the pre-trained model~\cite{ChenWFLW19,LiGD18,LiXWRLH19}. To mitigate the problem, \cite{LiGD18} constrains the distance between the fine-tuned model and its pre-trained counterpart. \cite{LiXWRLH19} proposes a data-dependent strategy that limits the distance between representations of target data before and after fine-tuning.

Besides catastrophic forgetting, another challenge is from the bias in pre-trained models. Due to the gap between learning tasks and data for pre-training and fine-tuning, the bias in the pre-trained model will degenerate the performance on the target task~\cite{QianHL20}. Without any side information, the bias is hard to be eliminated. Recent work~\cite{liu2021improved} shows that the bias can be reduced by fine-tuning the target data with a related subset selected from the pre-training data for the target task. In this work, we investigate another form of side information from text supervision. Compared with the information from pre-training data, the text supervision is easier to be accessed with a pre-trained text encoder.

\section{Visual Fine-tuning with Text Supervision}
Given an image data set of $\{x_i,y_i\}_{i=1}^n$, where $x_i$ denotes an image and $y_i$ is the corresponding label, a classification model can be learned by minimizing the empirical risk as
\begin{equation}\label{eq:scratch}
    \min_{\theta, W} \frac{1}{n}\sum_{i}^n \ell(\x_i,y_i)
\end{equation}
where $\x_i = f(x_i)\in\R^d$ is the representation extracted from a neural network $f(\cdot)$ and $\theta$ is the parameter of $f(\cdot)$. $\ell$ can be an appropriate loss function and cross entropy loss is prevalent for classification, which is also adopted in this work. The $C$-class classification loss can be written as
\begin{equation*}
    \ell(\x_i,y_i) = -\log\frac{\exp(\x_i^\top \w_{y_i})}{\sum_j^C \exp(\x_i^\top \w_j)}
\end{equation*}
where $W = [\w_1, \dots, \w_C]\in\R^{d\times C}$ is a linear classifier consisting of a single proxy from each class~\cite{QianSSHTLJ19}.

Due to the over-parameterization property of deep models~\cite{ZhuLS19}, i.e., the number of parameters in $\theta$ can be larger than that of training examples, directly optimizing the problem in Eqn.~\ref{eq:scratch} may incur the over-fitting problem, especially on data sets with a limited number of examples. To mitigate the problem, the model can be initialized by parameters pre-trained on a large-scale data set. Fine-tuning a pre-trained model with a small learning rate becomes prevalent for visual categorization~\cite{kornblith2019better}.

Compared with a randomly initialized model, the pre-trained model contains sufficient semantic information from the pre-training data that can be transferred to downstream tasks. However, the bias from the specific pre-training task and data can result in a sub-optimal performance when data distribution shifts. In this work, we propose to include the text supervision from the class names in the target task to help reduce the bias from the pre-trained model.

\subsection{Biased Classifier 
 in Conventional Fine-tuning}

First, we demonstrate that the bias in a pre-trained model will be preserved in the classifier $W$ after conventional fine-tuning. Moreover, even with a large learning rate solely for the classifier during fine-tuning, the obtained $W$ is still close to that implied by the pre-trained model.

Given fixed representations from a pre-trained model, the sub-problem of optimizing the classifier with cross entropy loss is convex and a global optimum can be obtained. Let $W^0=[\w_1^0,\dots,\w_C^0]$ be the optimal solution as
\begin{eqnarray}\label{eq:init}
W^0 = \arg\min_{W} \LL(\theta^0, W) = \frac{1}{n}\sum_i \ell(\x_i^{0}, y_i)
\end{eqnarray}
where $\x_i^0 = f_0(x_i)$ is extracted from the pre-trained model $f_0(\cdot)$, $\theta_0$ is the parameter of $f_0(\cdot)$, and $\w_j^0$ denotes the proxy for the $j$-th class obtained from the pre-trained model.

After fine-tuning $T$ iterations with stochastic gradient descent (SGD), we can observe $(\theta^T, W^T)$ for the problem
\begin{eqnarray}\label{eq:ft}
\min_{\theta,W:\|\theta - \theta^0\|_F\leq\epsilon} \LL(\theta,W) = \frac{1}{n}\sum_i \ell(\x_i, y_i)
\end{eqnarray}
where $W^T=[\w_1^T,\dots,\w_C^T)]$ and $\epsilon$ denotes the distance to the pre-trained backbone. A small $\epsilon$ is essential to avoid catastrophic forgetting in fine-tuning. We find that the proxies in $W^T$ are close to $W^0$ (i.e., the solution implied by the pre-trained model) as follows. All detailed proof of this work can be found in the appendix.

\begin{thm}\label{thm:1}
Let $W^0$ and $W^T$ be the optimal solution for Eqn.~\ref{eq:init} and that by SGD for Eqn.~\ref{eq:ft}. Assuming that $\LL(\theta^0, W)$ is $m$-strongly convex in $W$ and $\LL(\theta, W^T)$ is $L/2$-Lipschitz continuous in $\theta$, we have
\[\|W^T - W^0\|_F^2 \leq \frac{L}{m} \epsilon\]
\end{thm}

\paragraph{Remark~1} Theorem~\ref{thm:1} indicates that the changes in the classifier after fine-tuning heavily depends on the distance between backbones before and after fine-tuning, i.e., $\epsilon$. With a small learning rate for the backbone during the fine-tuning, the bias in the pre-trained model will be preserved, and thus can degenerate the performance on downstream tasks.

This can be further demonstrated by characterizing the gap $\epsilon$ between the pre-trained and fine-tuned models following the common practice in fine-tuning. During the fine-tuning, the backbone will be refined by an initial learning rate of $\eta_0$ with a cosine learning rate decay~\cite{kornblith2019better}. With SGD for optimization, $\epsilon$ can be bounded as follows.

\begin{prop}\label{prop:1}
Let $\theta^0$ and $\theta^T$ denote the parameters of the pre-trained backbone and those after fine-tuning. Assuming the gradient in SGD is bounded by $\|\nabla_{\theta}\LL\|_F\leq\delta$, we have
\[\|\theta^0 - \theta^T\|_F \leq 0.5\eta_0\pi\delta\]
\end{prop}

\paragraph{Remark~2} Proposition~\ref{prop:1} shows that with the cosine decay, the gap between the pre-trained and fine-tuned backbones mainly depends on the initial learning rate for the backbone. If adopting a small learning rate, the refined model will be close to the initial one, which can preserve the bias. However, increasing the learning rate is ineffective since it will lose the knowledge from the pre-trained model, and thus incur catastrophic forgetting. 

Incorporating the result in Proposition~\ref{prop:1}, the difference between proxies can be depicted as
\begin{cor}\label{cor:1}
Let $W^0$ and $W^T$ be the optimal solution for Eqn.~\ref{eq:init} and that by SGD for Eqn.~\ref{eq:ft}. With assumptions in Theorem~\ref{thm:1} and Proposition~\ref{prop:1}, we have
\[\|W^T - W^0\|_F^2 \leq \frac{\eta_0\pi\delta L}{2m}\]
\end{cor}

\paragraph{Remark~3} Corollary~\ref{cor:1} shows that the distance between classifiers depends on the initial learning rate for the backbone. Therefore, even with a different and larger learning rate for the classifier, it will still be close to that suggested by the pre-trained model.

Since it is hard to balance the trade-off for fine-tuning by the learning rate, we propose to include reference distributions from text supervision to mitigate the contradiction issue between bias and catastrophic forgetting.

\subsection{Reference Distribution from Text Supervision \label{optim}}

To reduce the potential bias existing in the pre-trained model, we introduce a reference distribution for the target task. Concretely, a distance constraint is added during the fine-tuning
\[\min_{\theta,W} \LL(\theta,W)\quad s.t. \quad D(W, W_r)\leq \alpha \]
where $W_r$ is a reference distribution of $W$ and $D(\cdot,\cdot)$ is a distance function, in which a squared Euclidean distance can be adopted as $D(W, W_r) = \|W-W_r\|_F^2$.

By constraining the distance to a reference distribution, the bias in the pre-trained model can be reduced effectively. In addition, the regularization is defined with proxies independent from the backbone, where a small learning rate is applicable for the backbone to avoid catastrophic forgetting.

However, obtaining an appropriate reference without any side information is challenging. Inspired by the recent progress in visual representation learning with natural language supervision~\cite{radford2021learning}, we consider leveraging the class name as text information to generate the reference distribution. Compared with images, tokens in text is organized discretely and text encoder can be pre-trained on a sufficiently large corpus with rich context, which is ideal to complement discrimination tasks such as classification. 

Let $Z = [\z_1,\dots,\z_C]\in\R^{d_z\times C}$ denote the proxies extracted from a text encoder for classes, the problem with text supervision can be written as
\[\min_{\theta,W} \LL(\theta,W)\quad s.t. \quad \|W - Z\|_F^2\leq \alpha\]
which is equivalent to
\[\min_{\theta,W} \LL(\theta,W) +\lambda \|W - Z\|_F^2\]

The Euclidean distance requires that the text feature has the same dimension as the vision feature, i.e., $d_z=d$, while these dimensions can vary with different encoders. By applying a projection function $h(\w):\R^{d}\to \R^{d_z}$, proxies of classes from different modalities can be manually aligned and the optimization problem becomes
\begin{eqnarray}\label{eq:align}
\min_{\theta, W} \LL(\theta,W) +\lambda \|h(W) - Z\|_F^2
\end{eqnarray}

However, due to the inherent differences between modalities, matching representations directly will introduce text-specific noise to visual tasks, which can degenerate the performance on the target visual task. In addition, compared to matching proxies of different modalities, the relationship between classes from the text encoder can be more informative for the target task. Therefore, we consider transferring the pairwise similarity between proxies from the text reference distribution to the vision task.

Concretely, given an anchor class $j$, the distribution over all classes can be computed
\begin{align}\label{eq:cdist}
&P_{j,k} = \frac{\exp(\tilde{\w}_j^{\top}\tilde{\w}_k/\tau)}{\sum_k^C\exp(\tilde{\w}_j^{\top}\tilde{\w}_k/\tau)}; P'_{j,k} = \frac{\exp(\tilde{\z}_j^{\top}\tilde{\z}_k/\tau')}{\sum_k^C\exp(\tilde{\z}_j^{\top}\tilde{\z}_k/\tau')}
\end{align}

where $P_j$ and $P'_j$ denote the distribution defined by proxies from the vision encoder and text encoder, respectively. $\tilde{\w}$ and $\tilde{\z}$ have the unit norm for $\w$ and $\z$, while $\tau$ and $\tau'$ are the temperature parameters for vision and text distributions, respectively. With the pairwise relations, we can constrain the KL-divergence between distributions as

\begin{eqnarray*}
\min_{\theta,W} \LL(\theta,W) +\lambda \sum_j^C D_{KL}(P'_j||P_j)
\end{eqnarray*}

To improve the efficiency of fine-tuning, the text encoder is fixed without fine-tuning. Therefore, representations of proxies $Z$ are only extracted once before fine-tuning the vision encoder. By eliminating the constant term from $P_j'$ in KL-divergence, the problem can be simplified with the cross entropy constraint as
\begin{eqnarray}\label{eq:ce}
\min_{\theta,W} \LL(\theta,W) - \lambda \sum_j^C \sum_k^C P'_{j,k}\log(P_{j,k})
\end{eqnarray}

Compared with the problem in Eqn.~\ref{eq:align}, KL-divergence focuses on the similarity between classes, and thus can reduce the noise from the modality-specific information.

\subsection{Instance-level Reference Distribution}

The class-level regularization in Eqn.~\ref{eq:ce} can be further improved by including the target task examples as a data-dependent regularization. First, if replacing the anchor class $j$ by an anchor example $x_i$, the revised instance-level distribution over vision classes becomes
\[P_{i,k} = \frac{\exp(\x_i^{\top}\w_k)}{\sum_k^C\exp(\x_i^{\top}\w_k)}\]
We can show that the class-level distribution is an approximation for the instance-level distribution and optimizing the instance-level distribution can help capture the variance in real data better.

\begin{thm}\label{thm:2}
Assuming that the norm of proxies is bounded by $\gamma$ as $\forall k, \|\w_k\|_2\leq \gamma$, the distribution defined by the anchor class is an approximation for the distribution defined by the anchor example as
\[\forall k, \quad \frac{1}{c^2} P_{y_i,k}\leq P_{i,k}\leq c^2 P_{y_i,k}\]
where $c = \exp(\gamma\|\x_i-\w_{y_i}\|_2)$.
\end{thm}

When the intra-class distribution is compact as $\|\x_i-\w_{y_i}\|_2\to 0$, the approximation becomes tight.

Unlike the class-level distribution, the instance-level distribution over text proxies is hard to compute due to the lack of corresponding text features for $x_i$. To mimic the data distribution in text space, we propose to optimize the cross entropy loss with the fixed text classifier for representation learning as
\[\ell_T(\x_i,y_i) = -\log\frac{\exp(h'(\x_i)^\top \tilde{\z}_{y_i}/\tau')}{\sum_j \exp(h'(\x_i)^\top \tilde{\z}_{j}/\tau')}\]
where $h'(\x):\R^{d} \to \R^{d_z}$ is the projection head with the unit normalization to project vision representations to the text space. Compared with Eqn.~\ref{eq:align}, the text-specific information will be learned for projected examples, which will not introduce noisy patterns to vision proxies.

With the approximated text representations $\x'_i = h'(\x_i)$, the distribution over fixed text proxies can be computed as
\[P'_{i,k} = \frac{\exp(\x_i^{'\top}\tilde{\z}_k/\tau')}{\sum_k^C\exp(\x_i^{'\top}\tilde{\z}_k/\tau')}\]

By optimizing the vision encoder and text projection simultaneously, the final objective of our proposed \textbf{Te}xt \textbf{S}upervised fine-tuning (TeS) can be cast as
\begin{align}\label{eq:obj}
&\min_{\theta, h', W} \underbrace{(1-\lambda_V)\LL(\theta, W) -  \frac{\lambda_V}{n}\sum_i^n \sum_k^C P'_{i,k}\log(P_{i,k})}_{\mbox{vision fine-tuning}} \nonumber\\
&+\underbrace{\frac{\lambda_T}{n}\sum_i \ell_T(\x_i,y_i)}_{\mbox{text projection}}
\end{align}
where the former two terms fine-tune the network for the target vision task and the latter one is for the projection head to obtain approximated distribution from the text space. During the inference, the projection head will be discarded and only the vision network is adopted for evaluation.

\begin{table*}[!ht]
	\centering
	\renewcommand{\arraystretch}{1.2}
	\setlength{\tabcolsep}{5pt}
	\scalebox{0.9}{
	\begin{tabular}{l c c c c c c c c c c c c}
		\hline
  \rotatebox{0}{Method}&\rotatebox{0}{Aircraft}&\rotatebox{0}{Caltech}&\rotatebox{0}{Cars}&\rotatebox{0}{C10}&\rotatebox{0}{C100}&\rotatebox{0}{CUB}&\rotatebox{0}{DTD}&\rotatebox{0}{Flower}&\rotatebox{0}{Food}&\rotatebox{0}{Pet}&\rotatebox{0}{SUN}&\rotatebox{0}{Avg.}\\
        \hline
        
        CE&87.70&91.39&89.98&97.75&84.50&76.63&73.19&96.39&87.70&90.13&59.03&84.95\\
        CE + LS&87.10&91.41&90.95&97.44&84.64&76.18&73.67&96.00&87.18&90.95&59.56&85.01\\
        CE + TLS&87.19&91.42&89.09&97.80&84.66&76.22&72.98&96.49&87.16&90.29&59.54&84.80\\
        \hline
        TeS + BERT&87.84&93.19&90.92&97.71&85.53&77.62&73.94&97.26&87.63&90.55&60.31&85.68\\
        
        \multirow{2}{*}{TeS + CLIP}&87.87&92.86&91.84&98.00&86.95&78.41&75.00&97.15&87.90&91.39&61.86&86.29\\
        &\textcolor{blue}{(+0.17)}&\textcolor{blue}{(+1.44)}&\textcolor{blue}{(+0.89)}&\textcolor{blue}{(+0.20)}&\textcolor{blue}{(+2.29)}&\textcolor{blue}{(+1.78)}&\textcolor{blue}{(+1.33)}&\textcolor{blue}{(+0.66)}&\textcolor{blue}{(+0.20)}&\textcolor{blue}{(+0.44)}&\textcolor{blue}{(+2.30)}&\textcolor{blue}{(+1.28)}\\
        \hline
	\end{tabular}
	}
    \vspace{1.5mm}
    \caption{Comparison with ResNet-50 pre-trained by MoCo-v2. ``LS'' and ``TLS'' denote conventional label smoothing and text guided label smoothing in Eqn.~\ref{eq:cdist}, respectively. The last row shows the accuracy improvement over the best baseline.}
	\label{ta:r50}
\end{table*}

\begin{table*}[!ht]
	\centering
	\renewcommand{\arraystretch}{1.2}
	\setlength{\tabcolsep}{5pt}
	\scalebox{0.9}{
	\begin{tabular}{l c c c c c c c c c c c c}
		\hline
  \rotatebox{0}{Method}&\rotatebox{0}{Aircraft}&\rotatebox{0}{Caltech}&\rotatebox{0}{Cars}&\rotatebox{0}{C10}&\rotatebox{0}{C100}&\rotatebox{0}{CUB}&\rotatebox{0}{DTD}&\rotatebox{0}{Flower}&\rotatebox{0}{Food}&\rotatebox{0}{Pet}&\rotatebox{0}{SUN}&\rotatebox{0}{Avg.}\\
        \hline
        CE&79.26&92.36&86.52&96.71&83.80&78.06&71.54&96.28&83.10&91.37&59.84&83.53\\
        CE+LS&79.11&92.34&86.69&96.23&83.42&78.86&71.65&96.25&82.79&91.76&59.59&83.52\\
        CE+TLS&79.41&91.73&86.84&96.53&82.93&78.17&71.60&95.89&82.44&91.83&59.67&83.37\\
        \hline
        \multirow{2}{*}{TeS + CLIP}&79.42&92.57&88.66&96.94&84.09&79.27&73.09&96.62&83.68&91.86&60.34&84.23\\
        &\textcolor{blue}{(+0.01)}&\textcolor{blue}{(+0.21)}&\textcolor{blue}{(+1.82)}&\textcolor{blue}{(+0.23)}&\textcolor{blue}{(+0.29)}&\textcolor{blue}{(+0.41)}&\textcolor{blue}{(+1.44)}&\textcolor{blue}{(+0.34)}&\textcolor{blue}{(+0.58)}&\textcolor{blue}{(+0.03)}&\textcolor{blue}{(+0.50)}&\textcolor{blue}{(+0.70)}\\
        \hline
	\end{tabular}
	}
    \vspace{1.5mm}
    \caption{Comparison with supervised pre-trained ResNet-18. The last row shows the accuracy improvement over the best baseline.}
	\label{ta:r18}
\end{table*}

\paragraph{Connection to label smoothing} While conventional label smoothing assigns the uniform weight for unrelated classes~\cite{abs-2006-11653}, our method can be considered as smoothing labels for each example with the reference distribution from the text classifier. The instance-level label smoothing is more flexible to model the diversity in target task data.

\section{Experiments}

To evaluate the proposed method, we conduct experiments on 11 downstream tasks including Caltech-101~\cite{fei2004learning}, CIFAR-10~\cite{krizhevsky2009learning}, CIFAR-100~\cite{krizhevsky2009learning}, Caltech-UCSD Birds 200 (CUB)~\cite{WahCUB_200_2011}, Describable Textures Dataset (DTD)~\cite{cimpoi2014describing}, FGVC Aircraft (Aircraft)~\cite{maji2013fine}, Food-101~\cite{bossard14}, Oxford-IIIT Pet (Pets)~\cite{parkhi2012cats}, Oxford 102 Flower (Flowers)~\cite{nilsback2008automated}, Stanford Cars (Cars)~\cite{krause2013collecting}, and SUN397~\cite{xiao2010sun}. Following the common practice~\cite{ding2022prompt,radford2021learning,zhou2022conditional,zhou2022learning}, we report mean per-class accuracy on Aircraft, Caltech, Flowers, and Pets, while Top-1 accuracy is reported for others.

Considering that label smoothing is closely related, two variants of label smoothing strategy are included in the comparison. First, the one-hot label in cross entropy loss can be smoothed by assigning uniform weights to unrelated classes~\cite{szegedy2016rethinking}, which is referred as ``CE+LS''. Besides, the uniform weight can be replaced by the distribution implied by a text encoder pre-trained by CLIP~\cite{radford2021learning} as in Eqn.~\ref{eq:cdist}, which is denoted as ``CE+TLS''. Finally, fine-tuning the original cross entropy loss is included as ``CE''.

\subsection{Implementation Details}
For an extensive comparison, ResNet~\cite{he2016deep} and ViT~\cite{dosovitskiy2020image}, two prevalent but diverse visual backbones are included in experiments. For ResNet, ResNet-50 pre-trained by MoCo-v2~\cite{chen2020improved} and a supervised pre-trained ResNet-18 are adopted, while ViT-B/32 pre-trained by MAE~\cite{he2022masked} and CLIP~\cite{radford2021learning} are applied for fine-tuning. Each model is fine-tuned with SGD~\cite{bottou2007tradeoffs} for 100 epochs. The batch size is 256 and the momentum is 0.9. The learning rate is searched in a range of 7 logarithmically-spaced values between $10^{-4}$ and $10^{-1}$ on a validation set. Weight decay is optional and if it is applied, the value is searched with the same setting between $10^{-6}$ and $10^{-3}$. The standard augmentations, i.e., random crop and random horizontal flipping, are applied as in standard fine-tuning pipelines. The parameter $\lambda_V$ in our method is fixed as 0.1 and shared by baseline methods with label smoothing, while tuning it may further improve our performance. The temperature $\tau'=0.03$ and $\lambda_T$ is searched in $[0.1,1.5]$ with the validation set. 

To obtain text supervision, two different language models are leveraged as the text encoder. First, we adopt the text encoder from CLIP~\cite{radford2021learning} that is pre-trained with a visual encoder. Besides, we also include BERT~\cite{devlin2018bert} to explore the differences between the pure language model and the multi-modal language model. To extract features for each class name, the same prompt in~\cite{radford2021learning}, i.e., ``a photo of a A'' or ``a photo of a A, a type of B'' for fine-grained data, is applied as input for text encoder. The projection function $h'$ consists of a 2-layer MLP suggested by the ablation experiment.

\subsection{Comparison on ResNet}
First, we compare different methods by fine-tuning a pre-trained ResNet. The comparison is mainly conducted on ResNet-50 as in Table~\ref{ta:r50}, while the results on ResNet-18 are summarized in Table~\ref{ta:r18}.

From Table~\ref{ta:r50}, we can observe that conventional label smoothing methods cannot improve the performance over the baseline with a distinct margin. It is because that these methods are developed to avoid over-fitting in pre-training, while fine-tuning can mitigate over-fitting by leveraging the pre-trained model. Consequently, the bias in the pre-trained model is more critical for fine-tuning, which cannot be addressed by existing label smoothing techniques. With the instance-level reference distribution from the text encoder, our method TeS can effectively improve the performance over all data sets. By applying the text encoder pre-trained by CLIP, TeS outperforms CE by $1.34\%$ in average with ResNet-50. Since the average performance with ResNet-50 already achieves about $85\%$, the improvement is relatively large. Finally, both text encoders can help fine-tune from ResNet-50 while the one from CLIP is $0.61\%$ better than BERT. It shows that the text encoder pre-trained with vision data is more appropriate for supervising the fine-tuning of vision tasks. Therefore, the text encoder from CLIP will be adopted in the following experiments.

By further examining Table~\ref{ta:r18}, we find that both supervised and unsupervised pre-trained models can benefit from TeS, which demonstrates that the proposed method is applicable for models pre-trained with different pretext tasks. 

\begin{table*}[!ht]
	\centering
	\renewcommand{\arraystretch}{1.2}
	\setlength{\tabcolsep}{5pt}
	\scalebox{0.9}{
	\begin{tabular}{l c c c c c c c c c c c c}
		\hline
  \rotatebox{0}{Method}&\rotatebox{0}{Aircraft}&\rotatebox{0}{Caltech}&\rotatebox{0}{Cars}&\rotatebox{0}{C10}&\rotatebox{0}{C100}&\rotatebox{0}{CUB}&\rotatebox{0}{DTD}&\rotatebox{0}{Flower}&\rotatebox{0}{Food}&\rotatebox{0}{Pet}&\rotatebox{0}{SUN}&\rotatebox{0}{Avg.}\\
        \hline
        CE&79.69&93.13&88.65&98.23&87.68&78.08&73.08&93.63&89.76&92.72&65.34&85.45\\
        CE + LS&78.82&93.65&88.67&98.15&87.50&78.98&74.26&93.02&89.96&92.98&66.76&85.71\\
        CE + TLS&78.36&93.99&88.64&98.20&87.79&80.31&75.05&93.48&90.01&93.05&66.72&85.96\\
        \hline
        \multirow{2}{*}{TeS + CLIP}&81.24&94.10&91.12&98.45&88.92&81.60&74.84&94.52&90.50&93.27&67.62&86.93\\
        &\textcolor{blue}{(+1.55)}&\textcolor{blue}{(+0.11)}&\textcolor{blue}{(+2.45)}&\textcolor{blue}{(+0.22)}&\textcolor{blue}{(+1.13)}&\textcolor{blue}{(+1.29)}&\textcolor{red}{(-0.21)}&\textcolor{blue}{(+0.89)}&\textcolor{blue}{(+0.49)}&\textcolor{blue}{(+0.22)}&\textcolor{blue}{(+0.86)}&\textcolor{blue}{(+0.97)}\\
        \hline
	\end{tabular}
	}
    \vspace{1.5mm}
    \caption{Comparison with ViT pre-trained by MAE. The last row shows the accuracy improvement over the best baseline.}
	\label{ta:mae}
\end{table*}

\begin{table*}[!ht]
	\centering
	\renewcommand{\arraystretch}{1.2}
	\setlength{\tabcolsep}{5pt}
	\scalebox{0.9}{
	\begin{tabular}{l c c c c c c c c c c c c}
		\hline
  \rotatebox{0}{Method}&\rotatebox{0}{Aircraft}&\rotatebox{0}{Caltech}&\rotatebox{0}{Cars}&\rotatebox{0}{C10}&\rotatebox{0}{C100}&\rotatebox{0}{CUB}&\rotatebox{0}{DTD}&\rotatebox{0}{Flower}&\rotatebox{0}{Food}&\rotatebox{0}{Pet}&\rotatebox{0}{SUN}&\rotatebox{0}{Avg.}\\
        \hline
        ZS&18.30&78.13&57.31&88.02&57.07&53.21&41.54&66.50&83.24&85.20&60.02&62.60\\
        CE&76.93&94.30&89.99&97.84&87.31&80.79&76.11&96.41&89.08&91.47&67.04&86.12\\
        CE + LS&77.28&94.78&89.25&97.98&88.79&78.82&76.33&96.32&88.42&91.57&70.28&86.35\\
        CE + TLS&77.99&93.62&89.37&97.91&87.91&78.27&76.49&95.73&87.89&92.11&70.05&86.12\\
        \hline
        \multirow{2}{*}{TeS + CLIP}&78.10&94.91&90.14&98.08&88.62&80.40&77.13&96.70&88.57&92.23&70.96&86.90\\
        &\textcolor{blue}{(+0.11)}&\textcolor{blue}{(+0.13)}&\textcolor{blue}{(+0.15)}&\textcolor{blue}{(+0.10)}&\textcolor{red}{(-0.17)}&\textcolor{red}{(-0.39)}&\textcolor{blue}{(+0.64)}&\textcolor{blue}{(+0.29)}&\textcolor{red}{(-0.51)}&\textcolor{blue}{(+0.12)}&\textcolor{blue}{(+0.68)}&\textcolor{blue}{(+0.55)}\\
        \hline
	\end{tabular}
	}
    \vspace{1.5mm}
    \caption{Comparison with ViT pre-trained by CLIP. The last row shows the accuracy improvement over the best baseline.}
	\label{ta:clip}
\end{table*}

\begin{table*}[!ht]
	\centering
	\renewcommand{\arraystretch}{1.2}
	\setlength{\tabcolsep}{5pt}
	\scalebox{0.9}{
	\begin{tabular}{l c c c c c c c c c c c c}
		\hline
  \rotatebox{0}{Method}&\rotatebox{0}{Aircraft}&\rotatebox{0}{Caltech}&\rotatebox{0}{Cars}&\rotatebox{0}{C10}&\rotatebox{0}{C100}&\rotatebox{0}{CUB}&\rotatebox{0}{DTD}&\rotatebox{0}{Flower}&\rotatebox{0}{Food}&\rotatebox{0}{Pet}&\rotatebox{0}{SUN}&\rotatebox{0}{Avg.}\\
        \hline
        CE&28.47&56.13&56.40&94.01&68.50&53.47&47.13&96.39&69.37&68.32&38.78&61.54\\
        CE+LS&28.80&57.73&58.75&93.75&68.63&53.78&48.38&96.48&68.22&63.17&39.55&61.57\\
        CE+TLS&29.92&56.62&58.85&94.31&67.09&53.24&47.21&96.10&65.63&64.92&39.52&61.22\\
        \hline
        \multirow{2}{*}{TeS + CLIP}&37.56&63.08&67.88&94.39&72.78&57.47&51.65&97.15&72.06&66.63&44.62&65.93\\
         &\textcolor{blue}{(+7.64)}&\textcolor{blue}{(+5.35)}&\textcolor{blue}{(+9.03)}&\textcolor{blue}{(+0.08)}&\textcolor{blue}{(+4.15)}&\textcolor{blue}{(+3.69)}&\textcolor{blue}{(+3.27)}&\textcolor{blue}{(+0.67)}&\textcolor{blue}{(+2.69)}&\textcolor{red}{(-1.69)}&\textcolor{blue}{(+5.07)}&\textcolor{blue}{(+4.36)}\\
        \hline
	\end{tabular}
	}
    \vspace{1.5mm}
    \caption{Comparison with 10\% randomly sampled training examples and ResNet-50 pre-trained by MoCo-v2. The last row shows the accuracy improvement over the best baseline.}
	\label{ta:10shot}
\end{table*}

\subsection{Comparison on ViT}
Then, we compare the proposed method with ViT as the pre-trained model. The provided projection head in CLIP vision encoder is kept as pre-trained parameters for projection. Table~\ref{ta:mae} and \ref{ta:clip} show the results with different pre-trained ViTs.

Evidently, a similar phenomenon as that with ResNet can be observed. First, the proposed TeS can improve the average performance for diverse downstream tasks, while conventional methods are ineffective for fine-tuning. Second, TeS surpasses CE by $1.48\%$ when fine-tuning the ViT pre-trained by MAE, which shows the effectiveness of our method for different architectures. Note that the vision encoder pre-trained by CLIP already incorporates the text side information in contrastive pre-training, which makes the gain from our method less than that by MAE. Nevertheless, TeS can still improve the average performance by $0.78\%$ over CE and the repeated experiments in appendix confirm that TeS outperforms the best baseline significantly. 

Third, fine-tuning vision encoders pre-trained by MAE and CLIP shows the similar average performance, while that for individual data sets varies. It demonstrates that different pre-training methods can encode diverse patterns, while the proposed method can obtain consistent improvement over various pre-trained models. Finally, compared with zero-shot transfer denoted as ``ZS'', fine-tuning outperforms ZS by a large margin of $24.3\%$ and it suggests that fine-tuning is preferred when training examples are sufficient.

\subsection{Comparison on Few-shot Learning}

In addition to the above comparison with fine-tuning the whole data set, the proposed method is also evaluated in a challenging scenario, where each data set only contains $10\%$ randomly sampled training examples for few-shot learning. The minimal number of examples will be set to 10 for each class. Pre-trained ResNet-50 is applied as the vision encoder and Table~\ref{ta:10shot} demonstrates the comparison.

With only 10\% training examples, it is challenging for fine-tuning to reduce the bias in the pre-trained model when distribution shifts. With text supervision, our method can explicitly constrain the learned distribution close to the target task and achieve $4.36\%$ improvement on average. While our method demonstrates the superior performance on fine-tuning the whole data set, the experiment on few-shot learning implies that it can be more helpful when the number of target training examples is limited.

\subsection{Comparison on Long-tailed Data}

Finally, we investigate the proposed method for the data set with long-tailed distribution. Following \cite{CaoWGAM19}, CIFAR-10 and CIFAR-100 with the imbalance ratio of $10$ and $100$ are included in the comparison. The performance of fine-tuning a pre-trained ResNet-50 is shown in Table~\ref{ta:imba}.

\begin{table}[!ht]
	\centering  
	\renewcommand{\arraystretch}{1.2}
	\setlength{\tabcolsep}{8pt}
	\scalebox{0.9}{
	\begin{tabular}{l p{0.6cm} p{0.6cm} p{0.6cm} p{0.6cm}}
	\hline
    Dataset&\multicolumn{2}{c}{CIFAR-10}&\multicolumn{2}{c}{CIFAR-100}\\
    \cmidrule(lr){2-3}
    \cmidrule(lr){4-5}
    Imba Ratio&\multicolumn{1}{c}{100}&\multicolumn{1}{c}{10}&\multicolumn{1}{c}{100}&\multicolumn{1}{c}{10}\\
    \hline
    CE&\multicolumn{1}{c}{88.71}&\multicolumn{1}{c}{95.71}&\multicolumn{1}{c}{54.95}&\multicolumn{1}{c}{75.55}\\
    TeS&\multicolumn{1}{c}{\textbf{89.80}}&\multicolumn{1}{c}{\textbf{96.00}}&\multicolumn{1}{c}{\textbf{57.76}}&\multicolumn{1}{c}{\textbf{77.69}}\\
    \hline
	\end{tabular}
    }
    \vspace{1.5mm}
    \caption{Comparison on long-tailed CIFAR-10 and CIFAR-100 by fine-tuning ResNet-50.}
	\label{ta:imba}
\end{table}

With the reference distribution from text, the distribution of proxies from minor classes will not be overwhelmed by that from major classes. Thus, the performance of TeS surpasses the baseline CE with a clear margin. The improvement becomes larger when the imbalance ratio increases to $100$ and it shows that our method is not sensitive for long-tailed data. 

To further demonstrate the proposed method for long-tailed data, we evaluate TeS on a challenging data set iNaturalist-2017~\cite{van2018inaturalist} that contains 5,089 categories in Table~\ref{ta:inat}. Evidently, our method outperforms cross entropy baseline with a large margin of $6.19\%$ and it confirms the potential of text supervision for handling long-tailed data.

\begin{table}[!ht]
	\centering
	\renewcommand{\arraystretch}{1.2}
	\setlength{\tabcolsep}{8pt}
	\scalebox{0.9}{
	\begin{tabular}{l c c}
    \hline
    Method&Acc\\
    \hline
    CE&42.37\\
    TeS&\textbf{48.56}\\
    \hline
	\end{tabular}
	}
    \vspace{1.5mm}
    \caption{Comparison of accuracy (\%) on iNaturalist-2017 with ResNet-50.}
	\label{ta:inat}
\end{table}

Since the proposed method is not optimized for class imbalance learning, combining it with state-of-the-art methods for this specific task~\cite{CaoWGAM19} is left as future work.

\subsection{Ablation Study}
In this subsection, we demonstrate the effect of different components in TeS by ablation study. Experiments are conducted with ResNet-50 on CIFAR-100.

\paragraph{\#Layer in Projection $h'$}
$h'$ contains multi-layer MLP to align vision representations with text classifier. We vary the number of layers and summarize the results in Table~\ref{ta:mlp}.

\begin{table}[!ht]
	\centering
	\renewcommand{\arraystretch}{1.2}
	\setlength{\tabcolsep}{5pt}
	\scalebox{0.9}{
	\begin{tabular}{l c c c c}
    \hline
    \#Layer&1&2&3&4\\
    \hline
    Acc\%&86.59&\textbf{86.95}&86.82&86.77\\
    \hline
	\end{tabular}
	}
    \vspace{1.5mm}
    \caption{Comparison of number of layers in projection function $h'$.}
	\label{ta:mlp}
\end{table}

When $h'$ has 1-layer MLP, it degenerates to a linear projection. If increasing the number of layers to $2$, the non-linear mapping helps improve the performance by $0.36\%$. However, a more complicated head will not further improve the performance, which is consistent with the observation in \cite{ChenK0H20}. Compared to ResNet-50, the computational overhead introduced by the 2-layer MLP is negligible, which keeps the efficiency of the proposed method.

\paragraph{Weight of Text Regularization $\lambda_T$}

$\lambda_T$ in Eqn.~\ref{eq:obj} weights the loss for projecting the vision representation to the text space spanned by the text classifier. We vary it in $\{0.1,0.3,0.5,0.7,0.9,1.1,1.3,1.5\}$ and Table~\ref{ta:lambda} summarizes the result.

\begin{table}[!ht]
	\centering
	\renewcommand{\arraystretch}{1.2}
	\setlength{\tabcolsep}{3pt}
	\scalebox{0.9}{
	\begin{tabular}{l c c c c c c c c}
    \hline
    $\lambda_T$&0.1&0.3&0.5&0.7&0.9&1.1&1.3&1.5\\
    \hline
    Acc\%&85.94&86.63&86.79&\textbf{86.95}&86.84&86.80&86.67&86.66\\
    \hline
	\end{tabular}
	}
    \caption{Comparison of $\lambda_T$ in Eqn.~\ref{eq:obj}.}
	\label{ta:lambda}
\end{table}

When $\lambda_T$ is small, the projection head cannot be learned sufficiently, which results in an inaccurate estimation for the reference distribution. By increasing $\lambda_T$, the projection head can approximate the text space effectively and a satisfied performance can be obtained.

\paragraph{Temperature for Text Classifier $\tau'$}

When approximating the representation in text space, a normalized Softmax operator is applied in cross entropy loss with a temperature parameter $\tau'$. We vary it in
\{0.01, 0.03, 0.05, 0.07, 0.1, 0.2, 0.3\} and Table~\ref{ta:tem} summarizes the result. It is obvious that a small temperature is preferred to obtain a sharp distribution for supervision. Note that we have the standard Softmax for vision encoder and there is no temperature for vision classifier.

\begin{table}[!ht]
	\centering
	\renewcommand{\arraystretch}{1.2}
	\setlength{\tabcolsep}{4pt}
	\scalebox{0.9}{
	\begin{tabular}{l c c c c c c c}
    \hline
    $\tau'$&0.01&0.03&0.05&0.07&0.1&0.2&0.3\\
    \hline
    Acc\%&86.56&\textbf{86.95}&86.83&86.28&85.99&85.70&85.27\\
    \hline
	\end{tabular}
	}
    \vspace{1.5mm}
    \caption{Comparison of the temperatures $\tau'$ in the text classifier.}
	\label{ta:tem}
\end{table}

\begin{table*}[!ht]
	\centering
	\renewcommand{\arraystretch}{1.2}
	\setlength{\tabcolsep}{8pt}
	\scalebox{0.9}{
	\begin{tabular}{l c c c c c c c c c c c c}
		\hline
  Method&Aircraft&Caltech&Cars&C10&C100&CUB&DTD&Flower&Food&Pet&SUN&Avg.\\
        \hline
        CE&87.70&91.39&89.98&97.75&84.50&76.63&73.19&96.39&87.70&90.13&59.03&84.95\\
        \hline
        TeS+CLIP-S&87.87&92.86&\textbf{91.84}&98.00&\textbf{86.95}&\textbf{78.41}&75.00&97.15&\textbf{87.90}&\textbf{91.39}&61.86&86.29\\
        TeS+CLIP-En&\textbf{88.23}&\textbf{93.18}&91.54&\textbf{98.08}&86.87&78.20&\textbf{75.32}&\textbf{97.18}&87.87&90.92&\textbf{61.99}&\textbf{86.31}\\
        \hline
	\end{tabular}
	}
    \vspace{1.5mm}
    \caption{Comparison of different prompt strategies on ResNet-50. ``CLIP-S'' and ``CLIP-En'' denote a single prompt and an ensemble of prompts for each class as in~\cite{radford2021learning}, respectively.}
	\label{ta:prompt}
\end{table*}

\begin{figure*}[!ht]
    \centering
    \includegraphics[width=0.95 \textwidth]{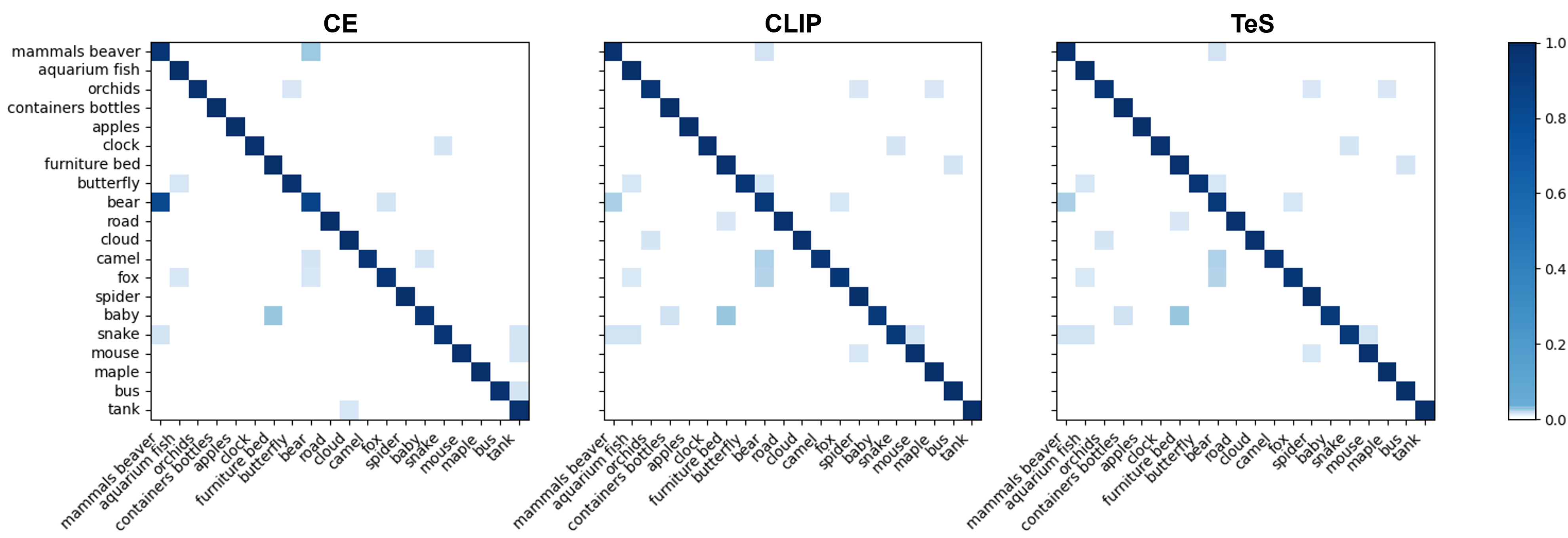}
    \caption{Illustration of confusion matrix from CE, CLIP, and TeS. Prediction on 20 classes from CIFAR-100 is shown for the comparison.}
    \label{fig:dist}
\end{figure*}

\paragraph{Prompt for Text Supervision}

Despite that a single prompt for each class demonstrates the superior performance with TeS, an ensemble of prompts shows better performance for zero-shot transfer in CLIP~\cite{radford2021learning}. Therefore, we also include the same ensemble strategy in TeS and the results are summarized in Table~\ref{ta:prompt}. While the ensemble is effective for zero-shot classification, the improvement for fine-tuning becomes marginal. Unlike the fixed pre-trained model in zero-shot learning, fine-tuning can leverage the labeled data to refine the pre-trained model. Hence, a single prompt is sufficient to generate appropriate text supervision for fine-tuning.

\paragraph{Variants of TeS}

Besides the instance-level distribution for regularization, the text supervision can be leveraged by directly mapping as in Eqn.~\ref{eq:align} and class-level distribution as in Eqn.~\ref{eq:ce}. We denote the two variants as TeS-M and TeS-C respectively and compare them to TeS in Table~\ref{ta:var}. In addition, the prediction from the text classifier is also evaluated.

\begin{table}[!ht]
	\centering
	\renewcommand{\arraystretch}{1.2}
	\setlength{\tabcolsep}{5pt}
	\scalebox{0.9}{
	\begin{tabular}{l c c c c c}
    \hline
    &CE&TeS-M&TeS-C&Text Head of TeS&TeS\\
    \hline
    Acc\%&84.50&84.78&84.94&86.20&\textbf{86.95}\\
    \hline
	\end{tabular}
	}
    \vspace{1.5mm}
    \caption{Comparison of variants for TeS. TeS-M denotes optimizing the objective in Eqn.~\ref{eq:align} while TeS-C minimizes the loss in Eqn.~\ref{eq:ce}.}
	\label{ta:var}
\end{table}

Compared with the baseline CE, all of these methods can improve the fine-tuning performance on CIFAR-100. Moreover, TeS-C outperforms TeS-M and it shows that pairwise alignment is more effective for regularization than the direct mapping. With instance-level supervision as in TeS, the accuracy can be further improved by more than $2\%$. Finally, the prediction from the text head is $0.75\%$ worse than that from the vision encoder, which demonstrates the modality gap between vision and language.

\paragraph{Illustration of Reference Distribution}

To illustrate the effect of reference distribution from text, we show the obtained confusion matrix from different methods in Fig.~\ref{fig:dist}. Note that there are 20 superclasses in CIFAR-100, 20 target classes from different superclasses are randomly sampled for comparison.

First, it is obvious that CE and CLIP have a different off-diagonal distribution, which shows the difference between pre-trained vision encoder and text encoder. Since text encoder can be pre-trained with a much larger corpus containing context information, its distribution can be leveraged as the reference for fine-tuning small data sets. Comparing TeS to CLIP, the off-diagonal patterns, e.g., the relation between ``baby'' and ``containers bottles'' and that between ``mouse'' and ``spider'', can be well preserved. On the contrary, many biased patterns in CE, e.g., ``orchids'' and ``butterfly'', ``bus'' and ``tank'' are removed in TeS, which demonstrates our proposal.

\section{Conclusion}
While leveraging pre-trained models for fine-tuning becomes prevalent for downstream tasks, reducing the bias in pre-trained models has been less investigated. In this work, we propose to apply class names as natural language supervision and obtain reference distribution to help adapt the fine-tuned model to the target data distribution. Experiments on diverse downstream classification tasks demonstrate the efficacy of text supervision. After the success on classification, applying our method for different vision tasks can be our future work.

\paragraph{Limit} To obtain the reference distribution, the exact name of each class is required as the input for the text encoder while it may be inaccessible due to the privacy policy. Therefore, the scenario without accurate class names can be challenging for the current method, which inspires a future research direction.

\paragraph{Acknowledgments} This work is supported by the Beijing Natural Science Foundation (No. JQ20023).

{\small
\bibliographystyle{ieee_fullname}
\bibliography{tes}
}

\appendix

\begin{table*}[!ht]
	\centering
	\renewcommand{\arraystretch}{1.2}
	\setlength{\tabcolsep}{5pt}
	\scalebox{0.9}{
	\begin{tabular}{l c c c c c c c c c c c c}
		\hline
  \rotatebox{0}{Method}&\rotatebox{0}{Aircraft}&\rotatebox{0}{Caltech}&\rotatebox{0}{Cars}&\rotatebox{0}{C10}&\rotatebox{0}{C100}&\rotatebox{0}{CUB}&\rotatebox{0}{DTD}&\rotatebox{0}{Flower}&\rotatebox{0}{Food}&\rotatebox{0}{Pet}&\rotatebox{0}{SUN}&\rotatebox{0}{Avg.}\\
        \hline

        CE + LS (mean)&76.80&94.76&89.21&98.02&88.59&78.79&75.95&96.12&88.29&91.57&69.92&86.17\\
        CE + LS (std)&0.46&0.17&0.12&0.08&0.16&0.08&0.09&0.50&0.31&0.04&0.34&0.21 \\
        \hline
        TeS 
 (mean)&\textbf{77.80}&94.78&\textbf{90.01}&97.97&88.48&\textbf{80.01}&\textbf{77.01}&\textbf{96.74}&88.49&\textbf{92.17}&\textbf{70.98}&\textbf{86.77}\\
        TeS (std)&0.16&0.10&0.10&0.11&0.10&0.32&0.12&0.10&0.08&0.13&0.11&0.13\\
        \hline
	\end{tabular}
	}
    \caption{Comparison with ViT pre-trained by CLIP. The significantly better method examined by Student's t-test is bolded.}
	\label{ta:clip}
\end{table*}

\section{Theoretical Analysis}

\subsection{Proof of Theorem~1}
\begin{proof}
Note that with the fixed features, the function $\LL(\theta^0, W)$ is convex in $W$. Assuming the function is $m$-strongly convex such that for the arbitrary $(W_1,W_2)$, we have
\[\LL(W_1)\geq \LL(W_2) + \langle\nabla_{W_2} \LL(W_2),W_1-W_2\rangle + \frac{m}{2}\|W_1-W_2\|_F^2\]
Since $W^0$ is the optimal solution for $\LL(\theta^0, W)$, we have
\begin{align}\label{eq:dist}
&\|W^T - W^0\|_F^2\leq \frac{2}{m}(\LL(\theta^0,W^T)-\LL(\theta^0,W^0))\nonumber\\
&= \frac{2}{m}(\LL(\theta^0, W^T) -\LL(\theta^T, W^T) + \LL(\theta^T,W^T) -\LL(\theta^0, W^0)) \nonumber\\
&\leq \frac{2}{m}(\LL(\theta^0,W^T) -\LL(\theta^T, W^T))
\end{align}
The last inequality is due to that fine-tuning can obtain a better performance than linear probing, i.e., $\LL(\theta^T,W^T)\leq \LL(\theta^0,W^0)$.

For fine-tuning, the loss function $\LL$ is non-convex but can be Lipschitz continuous. With $L/2$ as the parameter of Lipschitz continuous, we have
\begin{align*}
\LL(\theta^0,W^T) -\LL(\theta^T, W^T) \leq \frac{L}{2} \|\theta^0 - \theta^T\|_F\leq \frac{L}{2}\epsilon
\end{align*}
where the last inequality is from the constraint of fine-tuning. Taking it back to the Eqn.~\ref{eq:dist}, the result is obtained.
\end{proof}

\subsection{Proof of Proposition~1}

\begin{proof}
Note that the backbone is updated by SGD
\[\theta^t = \theta^{t-1} - \eta_{t} \nabla\LL_{\theta^{t-1}}\]
Adding $t$ from $0$ to $T$, we have $\theta^T = \theta^0 - \sum_t^T \eta_{t} \nabla\LL_{\theta^{t-1}}$. By applying the triangle inequality, the difference between $\theta^T$ and $\theta^0$ can be bounded as
\begin{align*}
&\|\theta^0-\theta^T\|_F = \|\sum_t^T \eta_{t} \nabla\LL_{\theta^{t-1}}\|_F \\
&\leq \sum_t^T \eta_{t}\|\nabla\LL_{\theta^{t-1}}\|_F\leq \sum_t^T \eta_{t}\delta
\end{align*}
With a cosine decay strategy and the initial learning rate as $\eta_0$, we have
\[\|\theta^0-\theta^*\|_F \leq 0.5\delta\eta_0\int_{0}^{\pi}1+cos(x)dx=0.5\eta_0\pi\delta\]
\end{proof}

\subsection{Proof of Theorem~2}
\begin{proof}
According to the definition, we have
\begin{align*}
&P_{i,k} = \frac{\exp((\x_i-\w_{y_i})^{\top}\w_k+\w_{y_i}^\top\w_k)}{\sum_j^C\exp((\x_i-\w_{y_i})^{\top}\w_j+\w_{y_i}^\top\w_j)}
\end{align*}
With Cauchy-Schwarz inequality, we have
\[-\gamma \|\x_i-\w_{y_i}\|_2\leq (\x_i-\w_{y_i})^{\top}\w_k\leq \gamma \|\x_i-\w_{y_i}\|_2\]
Due to the fact that exponential function is monotone, we have
\[P_{i,k}\leq \frac{c\exp(\w_{y_i}^\top\w_k)}{\sum_j^C\exp(\w_{y_i}^\top\w_j)/c} = c^2 P_{y_i,k}\]
and 
\[P_{i,k}\geq \frac{\exp(\w_{y_i}^\top\w_k)/c}{\sum_j^C c\exp(\w_{y_i}^\top\w_j)} = \frac{1}{c^2}P_{y_i,k}\]
where $c = \exp(\gamma\|\x_i-\w_{y_i}\|_2)$.
\end{proof}

\section{Repeated Experiments on CLIP}
We repeat experiments for the vision encoder of CLIP by 3 times and conduct Student's t-test at the $95\%$ confidence level in Table~\ref{ta:clip}. It confirms that our method is significantly better than the best baseline on average.

\end{document}